\pdfoutput=1
\documentclass[11pt,letterpaper]{article}
\usepackage{ijcnlp2017}
\usepackage{times}
\usepackage{latexsym}
\usepackage{times}
\usepackage{url}
\usepackage{graphicx}
\usepackage{amsmath,amssymb,amsthm}
\usepackage{microtype}
\usepackage{xcolor}
\usepackage{multirow}
\usepackage[linesnumbered,ruled]{algorithm2e}\SetInd{.4em}{.4em}
\usepackage{enumitem}
\usepackage{tikz}
\usetikzlibrary{shapes.multipart}
\DeclareMathOperator*{\argmax}{argmax}

\ijcnlpfinalcopy

\title{Improving Implicit Semantic Role Labeling \\by Predicting Semantic Frame Arguments}

  \author{Quynh Ngoc Thi Do$^1$, Steven Bethard$^2$, Marie-Francine Moens$^1$  \\
 $^1$Katholieke Universiteit Leuven, Belgium \\
  $^2$University of Arizona, United States  \\
 {\tt quynhngocthi.do@cs.kuleuven.be} \\
  {\tt bethard@email.arizona.edu} \\
  {\tt sien.moens@cs.kuleuven.be} \\}

\date{}
\begin{document}
\maketitle
\begin{abstract}
Implicit semantic role labeling (iSRL) is the task of predicting the semantic roles of a predicate that do not appear as explicit arguments, but rather regard common sense knowledge or are mentioned earlier in the discourse.
We introduce an approach to iSRL based on a predictive recurrent neural semantic frame model (PRNSFM) that uses a large unannotated corpus to learn the probability of a sequence of semantic arguments given a predicate.
We leverage the sequence probabilities predicted by the PRNSFM to estimate selectional preferences for predicates and their arguments.
On the NomBank iSRL
test set, our approach improves state-of-the-art performance on implicit semantic role labeling with less reliance than prior work on manually constructed language resources.
\end{abstract}

\section{Introduction}
Semantic role labeling (SRL) 
has traditionally focused on semantic frames
consisting of verbal or nominal predicates and \textit{explicit arguments} that occur \textit{within} the clause or sentence that contains the predicate.
However, many predicates, especially nominal ones, may bear arguments that are left implicit because they regard common sense knowledge or because they are mentioned earlier in a discourse \cite{Ruppenhofer:2010:STL:1859664.1859672,Gerber:2009:RIA:1620754.1620776}.
These arguments, called \textit{implicit arguments}, are resolved by another semantic task,
implicit semantic role labeling (iSRL).
Consider a NomBank \cite{meyers-EtAl:2004:HLTNAACL} annotation example: 
\begin{quote}
\textit{[$_{\textbf{A0}}$~The network] had been expected to have [$_{\textbf{NP}}$~losses] [$_{\textbf{A1}}$~of \$20 million] \ldots 
Those [$_{\textbf{NP}}$~losses] may widen because of the short Series.}
\end{quote}
The predicate \textit{loss} in the first sentence has two arguments annotated explicitly: A0, \textit{the entity losing something}, and  A1, \textit{the thing lost}. Meanwhile, the other instance of the same predicate in the second sentence has no associated arguments. However, for a good reader, a reasonable interpretation of the second \textit{loss} should be that it receives the same A0 and A1 as the first instance. These arguments are implicit to the second \textit{loss}.

As an emerging task, implicit semantic role labeling faces a lack of resources.
First, hand-crafted implicit role annotations for use as training data are seriously limited:
SemEval 2010 Task 10 \cite{Baker:1998:BFP:980845.980860} provided FrameNet-style \cite{Baker:1998:BFP:980845.980860} annotations for a fairly large number of predicates but with few annotations per predicate,
while \newcite{gerber:2010} provided PropBank-style \cite{Palmer-propbank} data with many more annotations per predicate but covering just 10 predicates.
Second, most existing iSRL systems depend on other systems (explicit semantic role labelers, named entity taggers, lexical resources, etc.), and as a result not only need iSRL annotations to train the iSRL system, but annotations or manually built resources for all of their sub-systems as well. 

We propose an iSRL approach that addresses these challenges, requiring no manually annotated iSRL data and only a single sub-system, an explicit semantic role labeler.
We introduce a predictive recurrent neural semantic frame model (PRNSFM), which can estimate the probability of a sequence of semantic arguments given a predicate, and can be trained on unannotated data drawn from the Wikipedia, Reuters, and Brown corpora, coupled with the predictions of the MATE \cite{bjorkelund-etal:2010:coling-demos} explicit semantic role labeler on these texts.
The PRNSFM forms the foundation for our iSRL system, where we use its probability estimates over sequences of semantic arguments to predict \textit{selectional preferences} for associating predicates with their implicit semantic roles.
Our PRNSFM-based iSRL model improves state-of-the-art performance, outperforming the only other system that depends on just an explicit semantic role labeler by 10 \% F1, and achieving equal or better F1 score than several other models that require many more lexical resources. 

Our work fits today's interest in natural language understanding, which is hampered by the fact that content in a discourse is often not expressed explicitly because it was mentioned earlier or because it regards common sense or world knowledge that resides in the mind of the communicator or the audience.
In contrast, humans easily combine relevant evidence to infer meaning, determine hidden meanings and make explicit what was left implicit in the text, using the \textit{anticipatory power} of the brain that predicts or ``imagines'' circumstantial situations and outcomes of actions \cite{Friston2010, Vernon2014} which makes language processing extremely effective and fast \cite{Kurby2015, Schacter2016}.
The neural semantic frame representations inferred by our PRNSFM take a first step towards encoding something like anticipatory power for natural language understanding systems.

The remainder of the paper is organized as follows: First, section \ref{relatedwork} describes the related work. Second, section \ref{lm} proposes the predictive recurrent neural semantic frame model including the formal definition, architecture, and an algorithm to extract selectional preferences from the trained model. Third, in section \ref{isrl}, we introduce the application of our PRNSFM in implicit semantic role labeling. Fourth, the experimental results and discussions are presented in section \ref{exp}. Finally, we conclude our work and suggest some future work in section \ref{conclusion}.


\section{Related work}\label{relatedwork}

\paragraph*{Language Modeling} Language models, from n-gram models to continuous space language models \cite{mikolov:embeddings,pennington2014glove}, provide probability distributions over sequences of words and have shown their usefulness in many natural language processing tasks. However, to our knowledge, they have not yet been used to model semantic frames. Recently, \newcite{DBLP:conf/acl/PengR16} developed two distinct models that capture semantic frame chains and discourse information while abstracting over the specific mentions of predicates and entities, but these models focus on discourse processing tasks, not semantic frame processing.

\paragraph*{Semantic Role Labeling} In unsupervised SRL, \newcite{woodsend-lapata:2015:EMNLP} and \newcite{titov-khoddam:2015} induce embeddings to represent a predicate and its arguments from unannotated texts, but in their approaches, the arguments are words only, not the semantic role labels, while in our models, both are considered.

\paragraph*{Low-resource Implicit Semantic Role Labeling}
Several approaches have attempted to address the lack of resources for training iSRL systems.
\newcite{impar} proposed an approach based on exploiting argument coherence over different instances of a predicate, which did not require any manual iSRL annotations but did require many other manually-constructed resources: an explicit SRL system, WordNet super-senses, a named entity tagger, and a manual categorization of SuperSenseTagger semantic classes. 
\newcite{Roth2015} generated additional training data for iSRL through comparable texts, but the resulting model performed below the previous state-of-the-art of \newcite{impar}.
\newcite{DBLP:conf/naacl/SchenkC16} proposed an approach to induce prototypical roles using distributed word representations, which required only an explicit SRL system and a large unannotated corpus, but their model performance was almost 10 points lower than the state-of-the-art of \newcite{impar}.
Similar to \newcite{DBLP:conf/naacl/SchenkC16}, our model requires only an explicit SRL system and a large unannotated corpus, but we take a very different approach to leveraging these, and as a result improve state-of-the-art performance.

\section{Predictive Recurrent Neural Semantic Frame Model} \label{lm}
Our goal is to use unlabeled data to acquire selectional preferences that characterize how likely a phrase is to be an argument of a semantic frame.
We rely on the fact that current explicit SRL systems achieve high performance on verbal predicates, and run a state-of-the-art explicit SRL system on unlabeled data.
We then construct a predictive recurrent neural semantic frame  model (PRNSFM) from these explicit frames and roles.

Our PRNSFM views semantic frames as a \textit{sequence}:
a predicate, followed by the arguments in their textual order, and terminated by a special EOS symbol. We draw predicates from PropBank verbal semantic frames, and represent arguments with their nominal/pronominal heads.
For example, \textit{Michael Phelps swam at the Olympics}
is represented as [\textit{swam}:PRED, \textit{Phelps}:A0, \textit{Olympics}:AM-LOC, EOS], where the predicate is labeled PRED and the arguments \textit{Phelps} and \textit{Olympics} are labeled A0 and AM-LOC, respectively.
Our PRNSFM's task is thus to take a predicate and zero or more arguments, and predict the next argument in the sequence, or EOS if no more arguments will follow.

We choose to model semantic frames as a sequence (rather than, say, a bag of arguments) because in English, there are often fairly strict constraints on the order in which arguments of a verb may appear.
A sequential model should thus be able to capture these constraints and use them to improve its probability estimates.
Moreover, a sequential model has the ability to learn the interaction between arguments in the same semantic frame.
For example, considering a swimming event, if \textit{Phelps} is A0, then \textit{Olympics} is more likely to be the AM-LOC than \textit{lake}.

Formally, for each $t^{th}$ argument of a semantic frame $f$, we denote its word (e.g., \textit{Phelps}) as $w_{f,t}$, its semantic label (e.g., A0) as $l_{f,t}$, where $w \in \bf V$, the word vocabulary, and $l \in {\bf L} \cup [\text{PRED}]$, the set of semantic labels.
We denote the predicate word and label, which are always at the $0^{th}$ position in the sequence, in the same way as arguments: $w_{f,0}$ and $l_{f,0}$.
We denote the sequence $[w_{f,0}, w_{f,1}, \ldots, w_{f,t-1}]$ as $w_{f,<t}$, and the sequence $[l_{f,0}, l_{f,1}, \ldots, l_{f,t-1}]$ as $l_{f,<t}$.
Our model aims to estimate the conditional probability of the occurrence of $w_{f,t}$ as semantic role $l_{f,t}$ given the preceding words and their labels:
\[P(w_{f,t}\text{:}l_{f,t}|w_{f,<t}\text{:}l_{f,<t})\]

We use a recurrent neural network to learn this probability distribution over sequences of semantic frame arguments.
For a semantic frame $f$ with $N$ arguments, at each time step $0 \leq t \leq N$, given the input $w_{f,t}\text{:}l_{f,t}$, the model computes the distribution $P(w_{f,t+1}\text{:}l_{f,t+1}|w_{f,<t+1}\text{:}l_{f,<t+1})$ and predicts the next most likely argument (or EOS).
During training, model parameters are optimized by minimizing prediction errors over all time steps.

We consider two versions of this model that differ in input (${\bf V_{in}}$) and output (${\bf V_{out}}$) vocabularies.

\subsection{Model 1: Joint Embedding LSTM}\label{model1}
\begin{figure*}
    \centering
\begin{tikzpicture}[
  baseline=\baselineskip,
  font=\small,
  hid/.style 2 args={
    rectangle split,
    rectangle split horizontal,
    draw=#2,
    rectangle split parts=#1,
    fill=#2!20,
    outer sep=1mm}]
  \pgfmathsetmacro{\height}{0.9}
  \pgfmathsetmacro{\width}{2}
  \pgfmathsetmacro{\last}{3}
  \node at (1, 2) {Output};
  \node at (1, 1) {Softmax};
  \node at (1, 0) {LSTM};
  \node at (1, -1) {Embedding};
  \node at (1, -2) {Input};
  \foreach \word/\label/\output [count=\step from 1] in {swam/PRED/Phelps:A0,Phelps/A0/Olympics:AM-LOC,Olympics/AM-LOC/EOS} {
    \node (i\step) at (2*\step*\width, -2*\height) {\strut\textit{\word}:\label};
    \node[hid={4}{purple}] (h\step) at (2*\step*\width, 0*\height) {};
    \node[hid={4}{blue}] (e\step) at (2*\step*\width, -1*\height) {};
    \node[hid={7}{brown}] (s\step) at (2*\step*\width, 1*\height) {};
    \node (o\step) at (2*\step*\width, 2*\height) {\strut\output};
  }
  \foreach \step in {1,...,\last} {
    \draw[->] (i\step.north) -> (e\step.south);
    \draw[->] (e\step.north) -> (h\step.south);
    \draw[->] (h\step.north) -> (s\step.south);
    \draw[->] (s\step.north) -> (o\step.south);
  }
  \foreach \step in {1,...,2} {
    \pgfmathtruncatemacro{\next}{add(\step,1)}
    \draw[->] (h\step.east) -> (h\next.west);
  }
\end{tikzpicture}
    \caption{Model 1 -- Joint Embedding LSTM}
    \label{fig:mdl1}
\end{figure*}
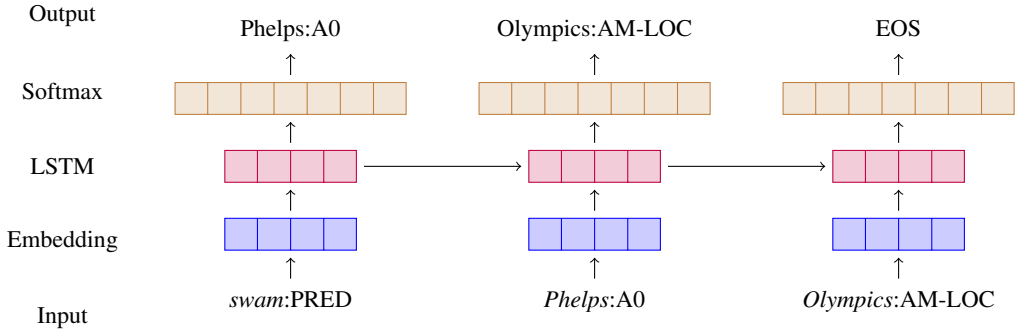
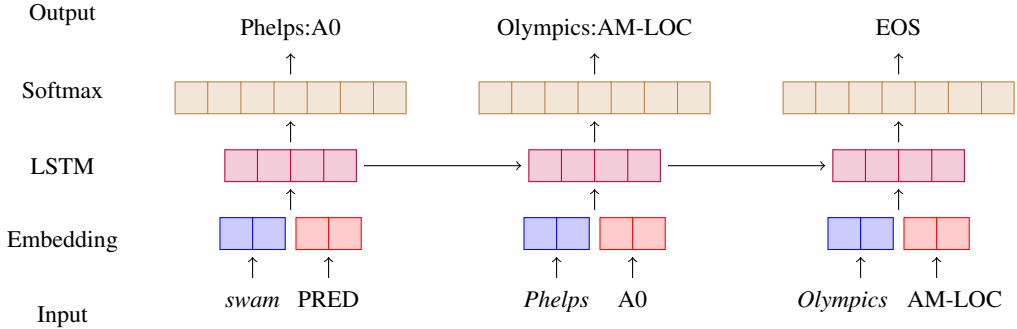
\begin{figure*}
    \centering
\begin{tikzpicture}[
  baseline=\baselineskip,
  font=\small,
  hid/.style 2 args={
    rectangle split,
    rectangle split horizontal,
    draw=#2,
    rectangle split parts=#1,
    fill=#2!20,
    outer sep=1mm}]
  \pgfmathsetmacro{\height}{0.9}
  \pgfmathsetmacro{\width}{2}
  \pgfmathsetmacro{\last}{3}
  \node at (1, 2) {Output};
  \node at (1, 1) {Softmax};
  \node at (1, 0) {LSTM};
  \node at (1, -1) {Embedding};
  \node at (1, -2) {Input};
  \foreach \word/\label/\output [count=\step from 1] in {swam/PRED/Phelps:A0,Phelps/A0/Olympics:AM-LOC,{Olympics\hspace*{1.5em}}/{\hspace*{1.5em}AM-LOC}/EOS} {
    \node (w\step) at (2*\step*\width - 0.25*\width, -2*\height) {\strut\textit{\word}};
    \node[hid={2}{blue}] (we\step) at (2*\step*\width - 0.25*\width, -1*\height) {};
    \node (l\step) at (2*\step*\width + 0.25*\width, -2*\height) {\strut\label};
    \node[hid={2}{red}] (le\step) at (2*\step*\width + 0.25*\width, -1*\height) {};
    \node[hid={4}{purple}] (h\step) at (2*\step*\width, 0*\height) {};
    \node[hid={7}{brown}] (s\step) at (2*\step*\width, 1*\height) {};
    \node (o\step) at (2*\step*\width, 2*\height) {\strut\output};
  }
  \foreach \step in {1,...,\last} {
    \draw[->] (w\step.north) -> (we\step.south);
    \draw[->] (l\step.north) -> (le\step.south);
    \draw[->] (e\step.north) -> (h\step.south);
    \draw[->] (h\step.north) -> (s\step.south);
    \draw[->] (s\step.north) -> (o\step.south);
  }
  \foreach \step in {1,...,2} {
    \pgfmathtruncatemacro{\next}{add(\step,1)}
    \draw[->] (h\step.east) -> (h\next.west);
  }
\end{tikzpicture}
    \caption{Model 2 -- Separate Embedding LSTM}
    \label{fig:mdl2}
\end{figure*}

We adopt the standard recurrent neural network language model \cite{conf/interspeech/MikolovKBCK10}, which is a natural architecture to deal with a sequence prediction problem. 

Model 1 consists of three layers (see Figure \ref{fig:mdl1}): an embedding layer that learns vector representations for input values; a long short-term memory (LSTM) layer that controls the sequential information receiving the vector representation as input; and a softmax layer to predict the next argument using the output of the LSTM layer as input.

This model treats the word and semantic label as a single unit in both input and output layers.
The model, therefore, learns joint embeddings for the word and its corresponding semantic label. 
For example, if we take ``Michael Phelps swam at the Olympics'' as training data, the three input values would be \textit{swam}:PRED, \textit{Phelps}:A0 and \textit{Olympics}:AM-LOC, and the three expected outputs would be \textit{Phelps}:A0, \textit{Olympics}:AM-LOC, EOS.
Since each word:label is considered as a single unit, the embedding layer will learn three vector representations, one for \textit{swam}:PRED, one for \textit{Phelps}:A0, and one for \textit{Olympics}:AM-LOC.
As can be seen, an important difference between our problem and the traditional language model is that we have to deal with two different types of information -- word and label.
By concatenating word and label, the standard recurrent neural network model can be applied directly to our data.


The detail of Model 1 is as following:

\paragraph{Embedding Layer} is a matrix of size $|{\bf V_{in}}| \times d$ that maps each unit of input into an $d$-dimensional vector. The matrix is initialized randomly and updated during network training. 

\paragraph{LSTM Layer} consists of $m$ standard LSTM units which take as input the output of the embedding layer, $x_{t}$, and produce an output $h_{t}$ by updating at every time step $0 \leq t \leq T$:
\begin{align*}
i_{t} &= sigmoid(W_ix_{t} + U_ih_{t-1}+b_i) \\
\hat{C_{t}} &= tanh(W_cx_{t} + U_ch_{t-1} + b_c)\\
f_{t}&=sigmoid(W_fx_{t} + U_fh_{t-1} + b_f)\\
C_{t}&=i_{t}*\hat{C_{t}} + f_{t} * C_{t-1}\\
o_{t} &= sigmoid(W_ox_{t} + U_oh_{t-1} + b_o)\\
h_{t}&=o_{t} * tanh(C_{t})
\end{align*}
where $W_i, W_c, W_f, W_o$ are weight matrices of size $d \times m$; $U_i, U_c, U_f, U_o$ are weight matrices of size $m \times m$; $b_i, b_c, b_f, b_o$ are bias vectors of size $m$; and $*$ is element-wise multiplication.
As per the standard LSTM formulation, $i_{t}$, $\hat{C_{t}}$, $f_{t}$, $C_{t}$, $o_{t}$  represent the input gate, states of the memory cells, activation of the memory cells' forget gates, memory cells' new state, and output gates' values, respectively.

\paragraph{Softmax Layer} computes the probability distribution of the next argument given the preceding arguments at time step $t$:
\begin{multline}
\label{eq:softmax}
P(w_{f,t+1}\text{:}l_{f,t+1}|w_{f,<t+1}\text{:}l_{f,<t+1}) =\\
softmax(h_{t}W + b) 
\end{multline}
where $W$ is a weight matrix of size $m \times |\bf V_{out}|$, and $b$ is a bias vector of size $|\bf V_{out}|$. The predicted next argument is:
\begin{align*}
\argmax_{w_{f,t+1}\text{:}l_{f,t+1}} P(w_{f,t+1}\text{:}l_{f,t+1}|w_{f,<t+1}\text{:}l_{f,<t+1}) 
\end{align*}
The network is trained using the negative log-likelihood loss function.

\subsection{Model 2: Separate Embedding LSTM}\label{model2}
Model 2 shares the same basic structure as Model 1, but considers the word and the semantic label as two different units in the input layer. 
As shown in Figure \ref{fig:mdl2}, we use two different embedding layers, one for word values and one for semantic labels, and the two embedding vectors are concatenated before being passed to the LSTM layer.
The LSTM and softmax layers are then the same as in Model 1.
For example, if we take ``Michael Phelps swam at the Olympics" as training data, the three input words would be \textit{swam}, \textit{Phelps}, and \textit{Olympics}, the three input roles would be PRED, A0 and AM-LOC, and the three expected outputs would be \textit{Phelps}:A0, \textit{Olympics}:AM-LOC, EOS.
A total of six different vector representations will be learned: a word embedding for each of \textit{swam}, \textit{Phelps}, and \textit{Olympics}, and a label embedding for each of PRED, A0 and AM-LOC.

In this model, the embedding layer for labels is initialized randomly (as in Model 1), but the embedding layer for word values is initialized with publicly available word embeddings that have been trained on a large dataset \cite{mikolov:embeddings}.

As compared to the joint-embedding Model 1, the separate-embedding Model 2 gives up a little power to represent the interaction between words and labels, but has a less sparse input vocabulary and gains the ability to incorporate pre-trained word embeddings.

\begin{figure*}
    \centering
    \includegraphics[width=1.5\columnwidth]{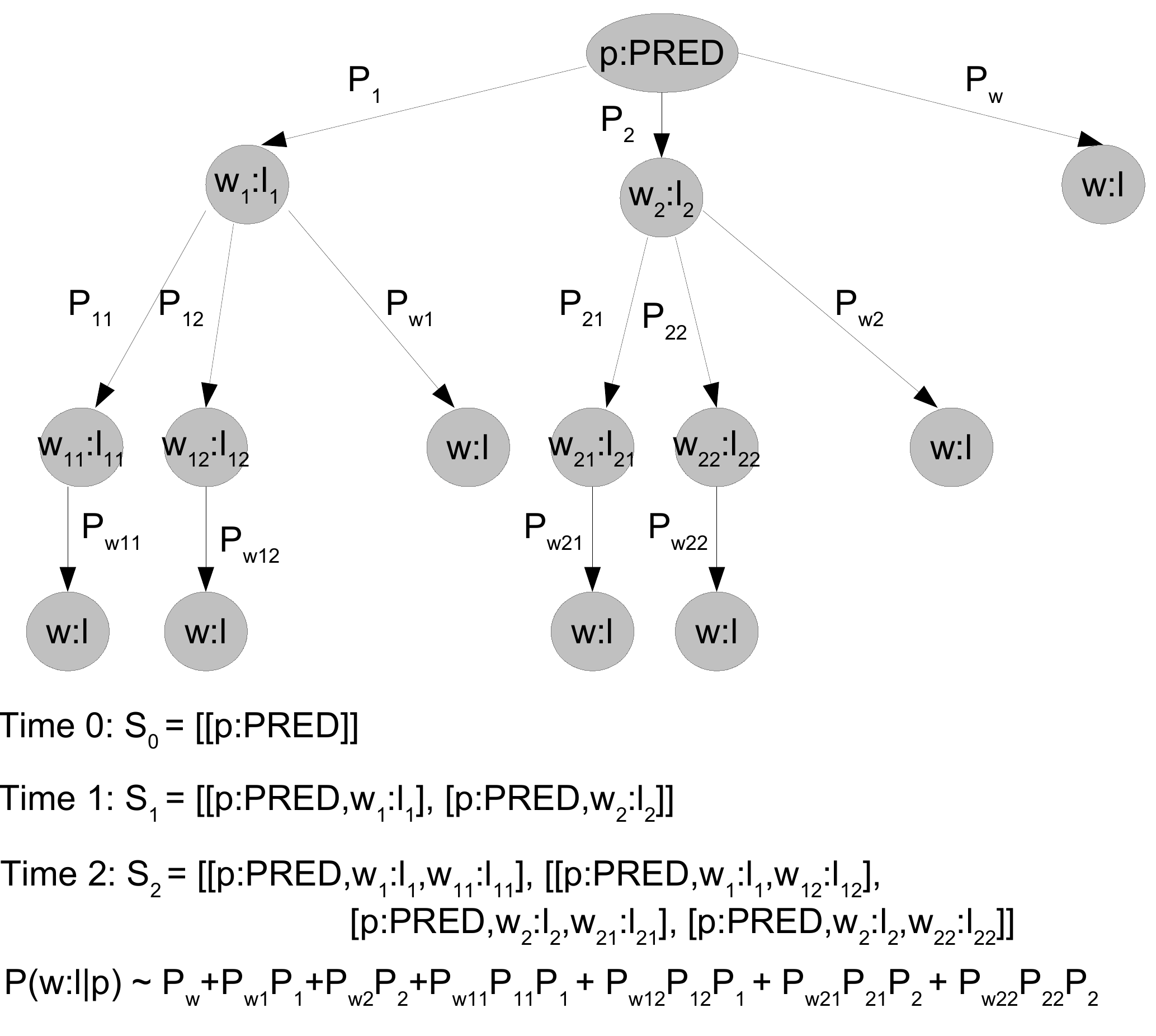}
    \caption{Selectional Preference  Inference example: $k$=2, $T$=3. The possible sequences are represented as a tree. Each arrow label is the probability of the target node to be predicted given the path from the tree root to the parent of the target node.}
    \label{fig:infe}
\end{figure*}

\subsection{Selectional Preferences}\label{selectionalpreference}
While the PRNSFM can predict the probability of an argument given the predicate and the preceding arguments, $P(w_{f,t}\text{:}l_{f,t}|w_{f,<t}\text{:}l_{f,<t})$, an iSRL system needs a \textit{selectional preference} score representing the probability of a word $w$ being the $l$ argument of predicate $p$, $P(w\text{:}l | p\text{:}PRED)$.
Thus, to convert our PRNSFM probabilities to selectional preferences, we need to marginalize over the possible argument sequences.

We approximate this marginalization by constructing a tree where the root is the predicate, $p$, the branches are likely sequences of arguments, and the leaves are the word and label for which we need to estimate a probability, $w\text{:}l$.
Formally, we define this tree of possible sequences as: 
\begin{equation*}
{\bf S_t} = \begin{cases}
\{[p\text{:}PRED]\} & \text{if } t=0\\
\begin{aligned}[t]
\{[&q, w_t\text{:}l_t]\colon q\in {\bf S_{t-1}}, \\
&w_t\text{:}l_t \in {\bf argmax}^{k}(q)
\}\end{aligned} & \text{if } 0<t<T\\
\{[q, w\text{:}l]\colon q\in {\bf S_{t-1}}\} & \text{if } t=T
\end{cases}
\end{equation*}
where $w_{f,0}\text{:}l_{f\text{:}0}=p\text{:}PRED$; $k$ and $T$ are thresholds; and ${\bf argmax}^{k}(q)$ is the $k$ word:label pairs that have the highest probability of being the next argument given the sequence q according to the PRNSFM.

We then estimate $P(w\text{:}l | p\text{:}PRED)$ as the sum of the probabilities of all the sequences encoded in the tree.
Formally:
\begin{align*}
P(w\text{:}l | p\text{:}PRED) &\approx \sum_{0 \leq t \leq T}{P(w\text{:}l|w_{f,<t+1}\text{:}l_{f,<t+1})}\\
&\approx\sum_{0 \leq t \leq T}  \sum_{q \in \bf S_{t} } P(w\text{:}l|q)\times P(q)
\end{align*}
where the probability of an argument sequence $q$ is the product of the PRNSFM's estimates for each step in the sequence:  
\begin{align}
\label{eq:ps}
P(q) &= P(w_{t}\text{:}l_{t}|w_{t-1}\text{:}l_{t-1}, \ldots,  p\text{:}\text{PRED}) \nonumber\\
&\times P(w_{t-1}\text{:}l_{t-1}|w_{t-2}\text{:}l_{t-2}, \ldots,  p\text{:}\text{PRED}) \nonumber\\
&\times\ldots 
\times P(w_1\text{:}l_1|p\text{:}\text{PRED})
\end{align}
An example of the calculation of $P(w\text{:}l | p\text{:}PRED)$ is shown in Figure \ref{fig:infe}.

Intuitively, the tree enumerates all possible argument sequences that start with the predicate, have zero or more intervening arguments, and end with the word and label of interest, $w\text{:}l$.
The probability of $w\text{:}l$ given the predicate is the sum of the probabilities of all branches in this tree, i.e., of all possible sequences that end with $w\text{:}l$.
In reality, we do not have the computational power to explore all possible sequences, so we must limit the tree somehow.
Thus, we only ask the PRNSFM for its top $k$ predictions at each branch point, and we only explore sequences with a maximum length of $T$.

\section{Implicit Semantic Role Labeling}\label{isrl}


As you will recall from previous sections, implicit semantic role labeling is the task of identifying discourse-level arguments of a semantic frame, which are missed by standard semantic role labeling, which operates on individual sentences. For instance, in ``This house has a new owner. The sale was finalized 10 days ago.'', the semantic frame evoked by ``sale'' in the second sentence should receive ``the house'' as an implicit A1 semantic role. Humans easily resolve the object of the sale given the candidates (in our example: ``house'' and ``owner''), but for a machine this is more difficult unless it has knowledge on what the likely objects of a sale are.
This kind of knowledge of selectional preferences can be extracted from our trained PRNSFM.

The previous section described how to extract selectional preferences from our PRNSFM.
However, that model is trained on verbal predicates, and the test data that we use \cite{gerber:2010} contains nominal predicates.
Thus, for each triple of a nominal predicate $np$, a word candidate $w$, and a label $l$, we approximate the selectional preference score of $w$ being the implicit argument role $l$ of $np$ as:
\begin{align*}
P(w\text{:}l|np)= max_{p \in V(np)} P(w\text{:}l|p\text{:}\text{PRED})
\end{align*}
where $P(w\text{:}l|p)$ is the selectional preference score described in Section~\ref{selectionalpreference}, and $V(np)$ is set of verbal forms of $np$. Here, we use the NomBank lexicon to get verbs associated with each nominal predicate, and then find instances of those verbs in the explicit SRL training data.
For example, for the noun \textit{funds}, $V(\textit{funds}) = \{\textit{funds}, \textit{fund},  \textit{funding}, \textit{funded}\}$. 

We apply selectional preferences to iSRL following \cite{impar}.
For each nominal predicate $np$ and implicit label $l$, the current and previous two sentences are designated the \textit{context window}.
Each sentence in the context window is annotated with the explicit SRL system.
If any instances of $np$ or $V(np)$ in the text have an explicit argument of type $l$, we deterministically predict the closest such argument as the implicit $l$ argument of $np$.
Otherwise, we run the PRNSFM over each word in the context window, and select the word with the highest selectional preference score above a threshold $s$. If all the candidates' scores are less than $s$, the system leaves the missing argument unfilled.
We optimized this threshold on the development data, resulting in $s = 0.0003$. 

As in \newcite{impar}, 
we apply a \textit{sentence recency factor} to emphasize recent candidates.
The selectional preference score $x$ is updated as $x'=x-z+z \times \alpha^d$ where $d$ is the sentence distance, and $\alpha$ and $z$ are parameters.
We set $z=0.00005$ based on the development set and set $\alpha=0.5$ as in \cite{impar}.

\section{Experiments} \label{exp}
We evaluate the two PRNSFM models on the iSRL task.
The tools, resources, and settings we used are as follows:

\paragraph{Semantic Role Labeling}
We used the full pipeline from MATE (\url{https://code.google.com/archive/p/mate-tools/}) \cite{bjorkelund-etal:2010:coling-demos} as the explicit SRL system, retraining it on just the CoNLL 2009 training portion.

\paragraph{Unannotated Data} The unannotated data used in the experiments
was drawn from
Wikipedia (\url{http://corpus.byu.edu/wiki/}), Reuters (\url{http://about.reuters.com/researchandstandards/corpus/}), and Brown (\url{https://catalog.ldc.upenn.edu/ldc99t42}).

\paragraph{Dataset for PRNSFM} The first 15 milion short and medium (less than 100 words) sentences from the unannotated  data (described above) were annotated automatically by the explicit SRL system. The obtained annotations were then used together with the gold standard CoNLL 2009 SRL training data to train the PRNSFM. 

\paragraph{Neural network training and inference} Parameters were selected using 
the CoNLL 2009 development set. We set the dimensions of word and label embeddings in the PRNSFM to 50 and 16, respectively. The hidden sizes of LSTM layers are the same as their input sizes. Word embedding layers are initialized by Skip-gram embeddings learned by training the word2vec tool \cite{mikolov:embeddings} on the unannotated data.
Our models were trained for 120 epochs using the AdaDelta optimization algorithm \cite{DBLP:journals/corr/abs-1212-5701}.
For fast selectional preference computing, we set $k=1$ and $T=4$\footnote{We selected relatively small values for the parameters to reduce the training and prediction time. We tried some larger values of the parameters on a small dataset, but found that the small values reported in the article achieved similar results with faster processing times.}.

\paragraph{Evaluation} We follow the evaluation setting in \newcite{gerber:2010,impar,DBLP:conf/naacl/SchenkC16}\footnote{
Following \newcite{DBLP:conf/naacl/SchenkC16}, we do not perform the alternative evaluation of \newcite{DBLP:journals/coling/GerberC12} that evaluates systems on the iSRL training set, since the iSRL training set overlaps with the CoNLL 2009 explicit semantic role training set on which MATE is trained.}: the method is evaluated on the  evaluation portion of the nominal iSRL data by Dice coefficient metrics. For each missing argument position of a predicate instance, the system
is required to either (1) identify a single constituent that fills the missing argument
position or (2) make no prediction and leave the missing argument position unfilled. To
give partial credit for inexact argument boundaries, predictions are scored by using the Dice
coefficient, which is defined as follows:
\begin{equation*}
Dice(predicted,true)=\frac{2\,|predicted\cap true|}{|predicted|+|true|}
\end{equation*}
$Predicted$ contains the tokens that the model has identified as the filler of the implicit argument position.
$True$ is the set of tokens from a single annotated constituent that truely fill the missing
argument position. The model's prediction receives a score equal to the maximum Dice
overlap across any of the annotated fillers (AF)\footnote{For iSRL, one implicit role may receive more than one annotated filler across a coreference chain in the discourse.}:
\begin{multline*}
Score(predicted)= \\
\max_{true \in AF} Dice(predicted,true)
\end{multline*}
Precision is equal to the summed prediction scores divided by the number of argument positions filled by the model. Recall is equal to the summed prediction scores divided by the number of argument positions filled in the annotated data.
\subsection{Experimental Setup}
In the {\bf baseline mode}, instead of using the PRNSFM, we only use the deterministic prediction by the explicit SRL system. We refer to this mode as \textit{Baseline} in Table \ref{tbl:isrl}. 

In the {\bf main mode}, the joint embedding LSTM model (Model 1) and the separate embedding LSTM model (Model 2) were trained on the same dataset which is a combination of the automatic SRL annotations and the gold standard CoNLL 2009 training data as described in the previous section. We denote this mode as \textit{gold CoNLL 2009 + unlabeled} in Table \ref{tbl:isrl}. 

To evaluate how well the system acquires knowledge from unlabeled data, we also train the PRNSFM only on the gold standard CoNLL 2009 training data.  We denote this mode as \textit{CoNLL 2009} in Table \ref{tbl:isrl}. 

In order to compare the performance of our sequential model to a non-sequential model, we train a skip-gram neural language model on the same unlabeled and labeled data as the PRNSFM in the main mode. The skip-gram model treats the predicates and arguments as a bag of labeled words rather than a sequence. The $P(w\text{:}l|p)$ is computed at the output layer of the skip-gram model by considering $w\text{:}l$ as the context of $p$. We denote this mode as \textit{Skip-gram} in Table \ref{tbl:isrl}. 

\subsection{Results and Discussion}
\begin{table*}[ht]%
\centering
\setlength{\tabcolsep}{0.45em}
\begin{tabular}{@{} l l c c c c c c c @{}}
\hline \hline
\textbf{Method} & \textbf{PRNSFM training data} & \rotatebox{90}{\textbf{iSRL data}} & \rotatebox{90}{\textbf{SRL system }} & \rotatebox{90}{\textbf{WordNet}} & \rotatebox{90}{\textbf{NER system }} & \textbf{P} & \textbf{R} & \textbf{F1}\\ \hline
\newcite{gerber:2010} &  & \checkmark & \checkmark & \checkmark & & 44.5  & 40.4  &  42.3 \\
\newcite{impar} &  &  & \checkmark & \checkmark & \checkmark & 47.9  & 43.8  &  45.8\\
\newcite{DBLP:conf/naacl/SchenkC16} &  &  & \checkmark & & & 33.5  & 39.2 & 36.1  \\ \hline

Baseline & & & \checkmark & &  & 75.3  & 17.2 & 28.0  \\ \hline
Skip-gram  & gold CoNLL 2009 + unlabeled & & \checkmark & &  & 26.3  & 32.3 & 29.0  \\ \hline
Model 1: Joint Embedding & gold CoNLL 2009 + unlabeled & & \checkmark & &  & 48.0  & 38.2  & 42.6 \\
Model 2: Separate Embedding & gold CoNLL 2009 + unlabeled & & \checkmark & & & 52.6  & 41.0  & \textbf{46.1} \\
Model 1: Joint Embedding & gold CoNLL 2009 & & \checkmark & &  & 39.2  & 34.1  & 36.5 \\
Model 2: Separate Embedding & gold CoNLL 2009 & & \checkmark & & & 40.2  & 36.0  & 38.0 \\
\hline \hline
\end{tabular}
\caption{Implicit role labeling evaluation.}
\label{tbl:isrl}
\end{table*}

Table \ref{tbl:isrl} shows the prior state-of-the-art and the performance of the baseline, skip-gram and our PRNSFM-based methods.


Our Model 2 achieves the highest precision and F1 score.
This is notable because the first two models require many more language resources than just an explicit SRL system: \newcite{gerber:2010} use WordNet and manually annotated iSRL data, while \newcite{impar} use WordNet, named entity annotations, and manual semantic category mappings.
\newcite{DBLP:conf/naacl/SchenkC16}, like our approach, use only an explicit SRL system, but both our models strongly outperform their results.  We assume that the difference here is caused by our proposed neural semantic frame model (PRNSFM). \newcite{DBLP:conf/naacl/SchenkC16} measure the selectional preference of a predicate and a role as a cosine between a standard word2vec embedding for the candidate word, and the average of all word2vec embeddings for all words that appear in that role. Our algorithms are very different: we take a language modeling approach and leverage the sequence of semantic roles, we learn custom word/role embeddings tuned for SRL, and then marginalize over many possible argument sequences. We assume that 
the learned PRNSFM representations are better informed about semantic frames than simple word embeddings, which only capture knowledge of contextual words.

Table \ref{tbl:isrl} also shows that training on large unlabeled data results in a marked improvement compared to training on only the CoNLL 2009 labeled data, providing evidence that the models have acquired linguistic knowledge from the unlabeled data. Although the automatically annotated data used to train the PRNSFM can be noisy, using a large amount of data has smoothed out the noise.  

Moreover, the better performance of our models over the standard skip-gram neural language model proves the effectiveness of modeling semantic frames as sequential data. The intuition here is that explicit semantic arguments have typical orderings in which they occur, so a sequential model should be a good fit for this problem.
Modeling this sequential aspect of the problem is effective, but requires us to marginalize out positional information to compute selectional preferences, since implicit semantic arguments can occur anywhere in the discourse and do not have a typical position.

Among our two models, Model 2, which learns separate vector representations for words and semantic roles, is better than Model 1, which learns a single vector representation of each (word, semantic role) pair.
The separate representation of words and roles means that Model 2 can share information across multiple occurrences of a word even if the semantic roles of that word are different, and this model can use publicly available embeddings pre-trained from even larger unannotated corpora when initializing its embeddings.

\newcite{DBLP:journals/coling/GerberC12} report an inter-annotator agreement of 64.3\% using Cohen's kappa measure on the annotated NomBank-based iSRL data. This value is borderline between low and moderate agreement indicating the sheer complexity of the annotation task, and explaining the relatively low performance of the iSRL systems.

\begin{table}[!t]%
\centering
\setlength{\tabcolsep}{0.1cm}
\begin{tabular}{l c c c c c}
\hline \hline
\textbf{Predicate} & Baseline & 2010 & 2013& 2016  & 2017 \\ \hline
 sale &36.2 & 44.2 & 40.3& 37.2& \textbf{52.8 } \\
 price & 15.4 & 34.2  & \textbf{53.3}&  27.3& 29.0  \\
investor & 9.8 & 38.4 &41.2 & 33.9& \textbf{43.1}  \\
bid & 32.3 & 21.3 & \textbf{52.0}&  40.7&  35.5 \\
plan &38.5  & 64.7 & 40.7& 47.4& \textbf{76.8} \\
cost &34.8  &\textbf{62.9} &53.0 & 36.9&  44.4 \\
loss &52.6 & \textbf{83.3 }& 65.8 & 58.9& 72.8  \\
loan & 18.2  &  37.5& 22.2&  37.9&\textbf{38.6}   \\
investment & 0.0 &30.8  & \textbf{40.8} & 36.6 & 23.5  \\
fund & 0.0 & 15.4&  \textbf{44.4} &  37.5& 42.8  \\
\hline \hline
\end{tabular}
\caption{A comparison on F1 scores (\%). 2010: \cite{gerber:2010}, 2013: \cite{impar}, 2016: Best model from \cite{DBLP:conf/naacl/SchenkC16}, 2017: Our best model (Model 2).}
\label{tbl:isrl-cpr}
\end{table}

In Table \ref{tbl:isrl-cpr}, we compare the F1 scores over all the ten predicates of our Model 2 to other state-of-the-art systems \footnote{As an overly conservative estimate, we take a t-test over the 10 predicate-level F1 scores as can be seen in Table \ref{tbl:isrl-cpr}.
Comparing against Model 2, this yields p=0.28 for \newcite{gerber:2010}, p=0.46 for \newcite{impar}, and most importantly p=0.058 for \newcite{DBLP:conf/naacl/SchenkC16}.}. Our system obtains relatively high scores ($>$ 50\%) on three predicates including ``sale'', ``plan'' and ``loss''. These three are the most frequent predicates (among the 10 defined in the nominal iSRL dataset) in the CoNLL 2009 training data -- they occur 1016, 318 and 275 times in verbal forms, respectively. In contrast, irregular predicates such as ``bid'' or ``loan'' usually have low performance. This is possibly caused by the dependence of our PRNSFM  on the performance of the explicit semantic role labeling system on verbal predicates. 

It is important to consider how iSRL can be extended beyond the 10 annotated predicates of \newcite{gerber:2010}.
Our models do not require any handcrafted iSRL annotations for training, and thus can be applied to all predicates observed in large unannotated data on which they are trained. 

However, as other work in iSRL, our approach still relies on a resource-heavy SRL system to learn selectional preferences. It would be interesting to investigate in further studies whether this SRL system can be replaced by a low-resource system \cite{Collobert:2011:NLP:1953048.2078186,connor2012starting}.

\section{Conclusion and Future Work}\label{conclusion}
We have presented recurrent neural semantic frame models for learning probability distributions over semantic argument sequences.
By modeling selectional preferences from these probability distributions, we have improved state-of-the-art performance on the NomBank iSRL task while using fewer language resources.
In the future, we believe that our semantic frame models are valuable in many language processing tasks that require discourse-level understanding of language, such as summarization, question answering and machine translation. 

\section*{Acknowledgment}
This work is carried out in the frame of the EU CHIST-ERA project ``MUltimodal processing of Spatial and TEmporal expRessions'' (MUSTER), and the ``MAchine Reading of patient recordS'' project (MARS, KU Leuven, C22/015/016).
\bibliography{ijcnlp2017}
\bibliographystyle{ijcnlp2017}

\end{document}